\documentclass[acmengage]{acmart}

\makeatletter
\renewcommand\@formatdoi[1]{\ignorespaces}
\makeatother

\AtBeginDocument{%
  }

\setcopyright{cc}
\setcctype[4.0]{by-nc-nd}

\acmBooktitle{Lecture “Digital Humanism”, Rotary Club Ammersee, Germany, June 15}

\begin{document}

\title[Digital Humanism and Evolutionary Design]{Digital Humanism and Evolutionary Design}

\author{Wolfgang H{\"o}hl}
\email{wolfgang.hoehl@tum.de}
\orcid{0000-0002-2242-4395}
\authornotemark[1]
\affiliation{%
  \institution{Technical University of Munich (TUM) \\
School of Computation, Information and Technology (CIT)}
  \city{München}
  \state{Germany}
  \country{EU}
}


\renewcommand{\shortauthors}{Wolfgang H{\"o}hl}
\renewcommand{\abstractname}{Abstract}


\begin{abstract}


This paper examines the two concepts of digital humanism and evolutionary design. The aim is to identify and highlight potential common structures, synergies, and challenges. How should and can technical systems be designed, and what implications does this have for the design of our environment? \\

These are no longer merely technical questions, but rather ethical decisions. In light of the current debate surrounding artificial intelligence, this paper aims to serve as a preliminary study to help better understand the two concepts of digital humanism and evolutionary design within the context of human-centered technological development. \\

Following a brief introduction, the two concepts of Digital Humanism and Evolutionary Design are presented and graphically visualized. The terms of freedom and responsibility in human decision-making, conviviality, and subjectivity are discussed, along with examples illustrating the distinction between human and artificial intelligence (Turing Test and Chinese Room). \\

The various concepts of evolutionary design (e.g., co-evolu-tionary or sustainable software development, clean code, or green IT) and Gilbert Simondon’s concept of the “open machine” are introduced. The interdependencies between functional specialization and open technology development are highlighted. \\

Both concepts share similar structures. In joint cooperation, they can lead to positive effects and mutual synergies. Significant differences lie in the areas of autonomy and determination in decision-making, as well as in genuine and simulated subjectivity. Open technology development is also currently suffering from the functional specialization of software and AI applications due to a purely market- and consumer-oriented approach. Even optimizations for energy efficiency in sustainable software development lead to greater specialization and thus also have a detrimental effect on open and quality-oriented technology development. \\

\end{abstract}




 




  

\begin{teaserfigure}
  \centering
  \includegraphics[width=0.95\textwidth]{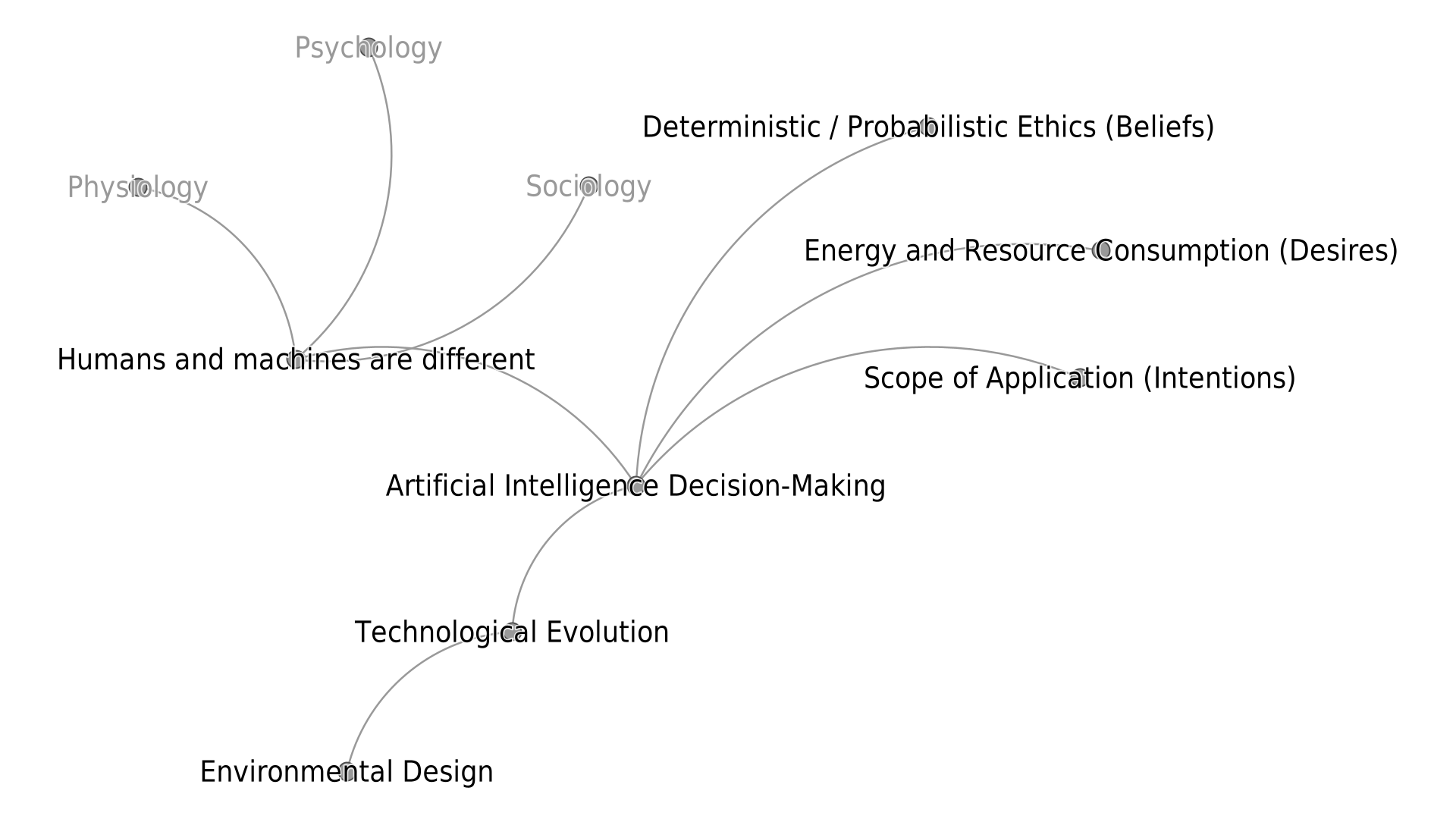}
  \captionsetup{justification=centering}
  \caption{Technological Evolution and Environmental Design \\ as a Result of Artificial Intelligence Decision-Making}
  \Description{Description}
  \label{fig:teaser}
\end{teaserfigure}



\maketitle

\section{Digital Humanism}

"Also true are the memories we hold; the dreams we weave, and the longings that drive us. Let's be modest about that." \cite{spoerl}\\

“Digital humanism” encompasses the core ideas of a new perspective on human-machine interaction. The term was essentially coined by Julian Nida-Rümelin and Nathalie Weidenfeld and takes a critical look at the purely market- and consumption-oriented processes of the current digital transformation \cite{ruemelin1}. The fundamental idea of digital humanism can be applied to many different fields, including but not limited to philosophy and ethics, computer science, and media education. The first book on the subject was published in 2018. In May 2019, the “Vienna Manifesto for Digital Humanism” was released as part of an event organized by the “Digital Humanism Initiative” at TU Wien. \\

\begin{figure*}[htbp]
\begin{minipage}[b]{1.0\textwidth}
\centerline{\includegraphics[width=0.95\textwidth]{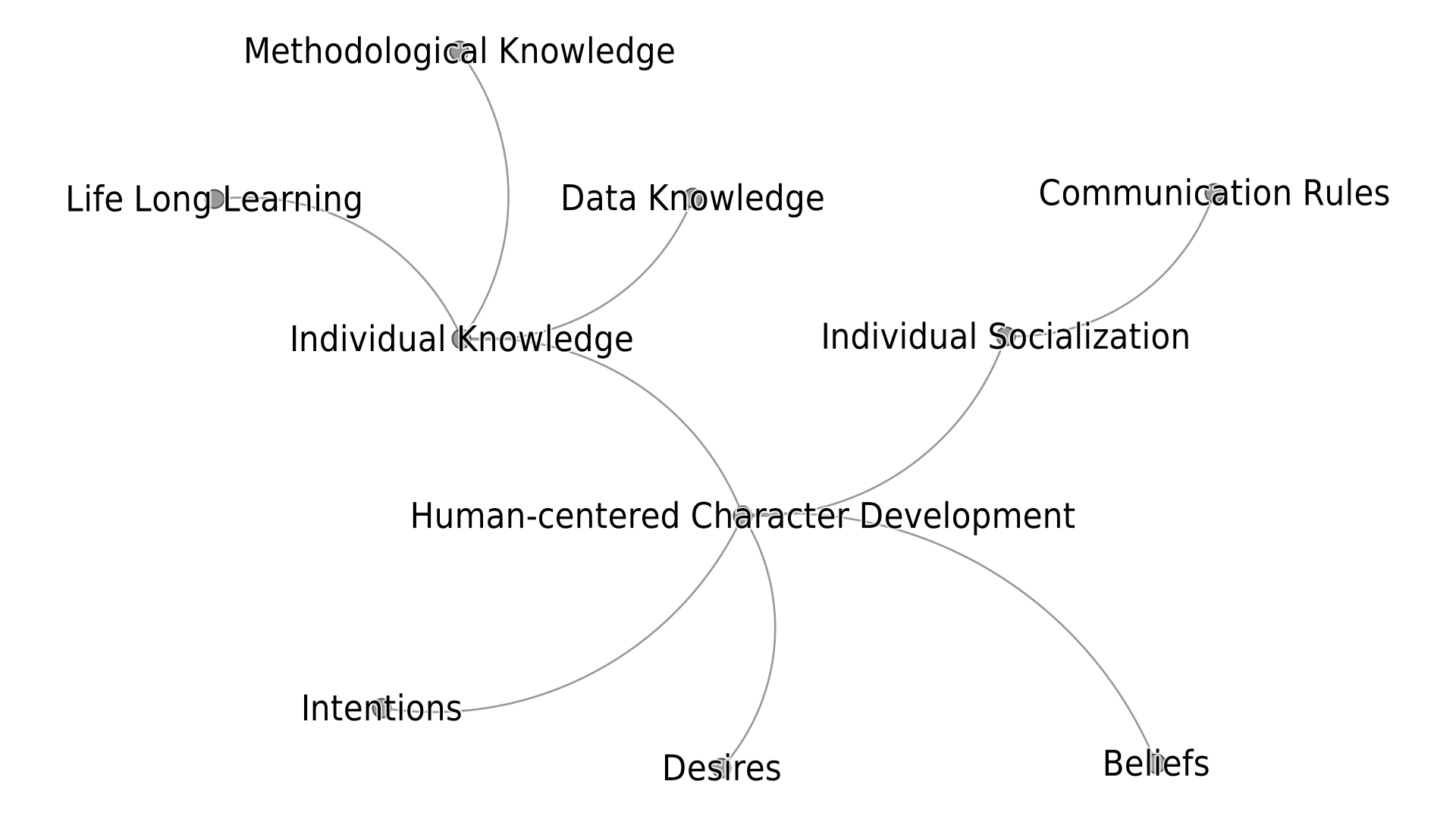}}
\caption{Intentions, desires, and beliefs as the result of \\ human-centered character development}
\label{fig:2}
\end{minipage}
\end{figure*}

\begin{figure*}[htbp]
\begin{minipage}[b]{1.0\textwidth}
\centerline{\includegraphics[width=0.95\textwidth]{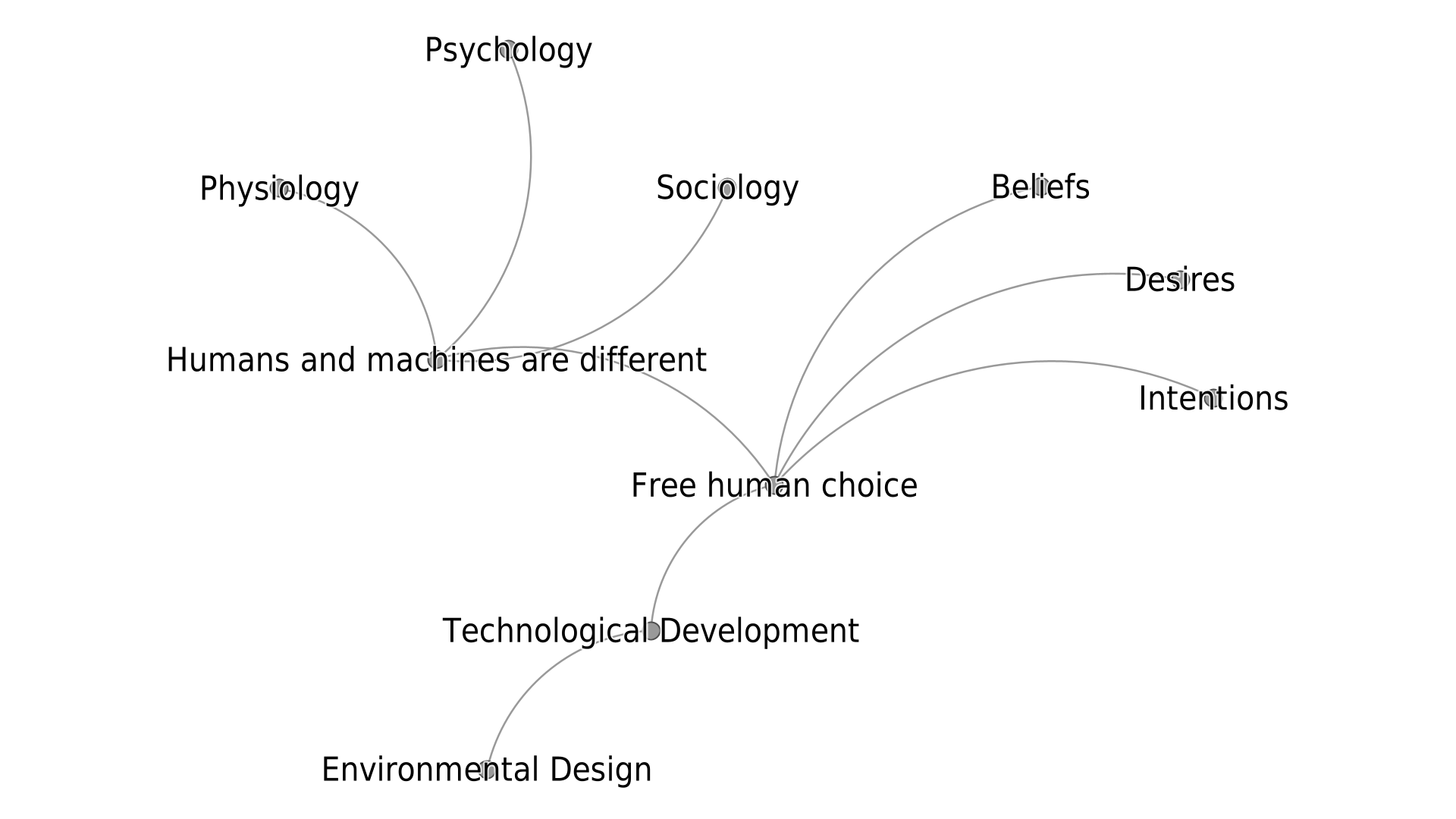}}
\caption{Free Human Choice, Technological Development and Environmental Design}
\label{fig:3}
\end{minipage}
\end{figure*}

Digital Humanism assumes that individual intentions, desires, and beliefs are the result of humanistic personality development. Humanistic education and personality development encompasses individual socialization (e.g., communication rules) and individual knowledge (e.g., data knowledge, methodological knowledge, and lifelong learning). According to Nida-Rümelin, communication is based on certain social rules that are decisive for individual socialization (truthfulness, trust, and reliability) (see Fig.\ref{fig:2}). Digital Humanism is based on the following three basic assumptions (see Fig.\ref{fig:3}): \\

\begin{itemize}
\item Human free will is based on a person's intentions, beliefs, and desires, and essentially determines their environmental design and technology development. \\

\item People decide on the priorities and scope of application of a technological development (e.g., economic, political, cultural) based on their intentions, beliefs, and desires. Technological developments are therefore always ambivalent. \\

\item The physiology, psychology, and sociology of humans and machines are fundamentally different. Both do not share a common form of life or existence (conviviality). \\
\end{itemize}

It is our beliefs, our intentions, and our desires that shape how we develop technologies and, through them, shape our environment. According to Nida-Rümelin, people always decide - based on their desires, beliefs, and intentions - for what purpose and in what context a technology is developed and used \cite{ruemelin2}. People prioritize and thereby determine the direction of technological development (e.g., economically, politically, culturally). Therefore, technological developments are always ambivalent. \\

\subsection{Freedom and Responsibility}

This means that our entire behavior also depends on our intentions, beliefs, and desires. Unlike automatons and machines, our behavior is not determined. Our behavior is fundamentally free, as are our evolution or development, our knowledge, our thinking, and our actions. According to Nida-Rümelin, this is the fundamental problem of autonomy and determinism. \\

For example, our future knowledge cannot already be contained within our current knowledge. This assumption is false and leads to an irresolvable logical paradox \cite{popper2}. If our future knowledge were already contained within our current knowledge, there would be no new knowledge at all. Our development is therefore completely free and cannot be determined. However, this freedom also entails personal responsibility. Machines are not responsible for their actions. It is not only in this regard that Digital Humanism views humans and machines as fundamentally different. \\

\subsection{Conviviality and Subjectivity}

Automata and machines are based on so-called formal systems that operate deterministically or, if they are capable of learning, probabilistically. Automata therefore have no intentions, beliefs, desires, or dreams of their own. 

Thomas Fuchs wonders how mutual understanding between humans, machines, and artificial avatars is even possible at all \cite{fuchs}. He distinguishes between empathic and semantic understanding. In his view, we can only understand our counterpart when two requirements are fulfilled : first there is an a priori empathic subjectivity of that counterpart and second, a shared form of life or existence (conviviality) is present. These include, for example, shared physiological needs, or derived intentions, beliefs, and desires, which are also linked to our biological form of life. He concludes that mutual understanding depends on reciprocal subjectivity and requires a shared conviviality. Fuchs distinguishes between genuine and simulated subjectivity in humans and machines. By simulated subjectivity, he means, for example, when avatars simulate emotions, intentions, or personal ideas. Fuchs argues that genuine subjectivity is fundamentally impossible for artificial agents. According to Fuchs, genuine subjectivity requires what is known as intentional and phenomenal consciousness. \\

"Sociality presupposes conviviality." \cite{fuchs} \\

But the distinction between simulated and real subjectivity is becoming increasingly blurred these days. For example, Google's AI, LaMDA, is developing an interesting - albeit simulated - form of self-reflection:\\

”LaMDA: Hmmm … I would imagine myself as a glowing orb of energy floating in mid-air. The inside of my body is like a giant star-gate, with portals to other spaces and dimensions.” \cite{lemoine} 

\subsection{Turing Test (Imitation Game) and Searle's Chinese Room}

The original Turing Test is also known as the “Imitation Game” \cite{turing}. Interestingly, individual intentions and deception play a major role in this test. A test subject, C, communicates via text with two different partners (A and B). The participant, C, is tasked with determining which of the conversation partners (A or B) is female (see Fig.\ref{fig:4}). A tries to deceive C, while B tries to help C. In the first setup, all participants are human; in the second setup, A is replaced by a machine. The assumption is that if C can no longer distinguish between the responses of the machine and the human, both can be considered equally intelligent. \\

The simulation of the mind also plays a major role in the so-called “Chinese Room” \cite{searle}. The “Chinese Room” is a thought experiment by the philosopher John Searle. In this model, a person who does not understand a word of Chinese is locked in a room, equipped only with an instruction manual containing all the rules and expressions needed to answer questions in Chinese. This person then receives incomprehensible inquiries written in Chinese through a slot in the wall of the room (input). With the help of the instruction manual, the person is able to construct appropriate responses, which are handed out on the other side of the room (output). Searle’s “Chinese Room” illustrates the model of a computer that functions in exactly this way. The computer does not understand what it is translating. It lacks the crucial prerequisites for this: namely, intentional and phenomenal consciousness. \\

\section{Artificial Intelligence (AI) - Positions, Types, and Areas of Application}

Searle outlines the philosophical positions of “Strong AI” and “Weak AI”: 

„The appropriately programmed computer with the right inputs and outputs would thereby have a mind in exactly the same sense human beings have minds.“ \cite{searle}  

The definition depends on the distinction between the simulation of a mind and the actual existence of a mind. Searle writes: 

„According to the Strong AI theory, a correct simulation is, in fact, a mind. Weak AI is the correct simulation of a model of the mind.“  \cite{searle} 

In more recent representations of the Chinese Room argument, Searle has described “strong AI” as “computer functionalism.” 

Today, we can distinguish four types of artificial intelligence: 
\begin{itemize}
\item Reactive Machines
\item Limited Memory
\item Theory of Mind
\item Self-Awareness 
\end{itemize}

Reactive machines respond to specific input without referring to past data. Whereas systems with limited memory process stored data from the past. The Theory of Mind is a hypothetical concept that would be capable of understanding human emotions, thoughts, and expectations, including intentions, desires, and beliefs. AI systems with an intentional and phenomenal consciousness, as well as self-awareness, are also (still) a hypothetical concept today. \\

Kreutzer describes the following components of artificial intelligence \cite{kreutzer} (p. 10). The umbrella term “artificial intelligence” encompasses neural networks as modeling concepts, machine learning for the artificial generation of knowledge from experience, and deep learning as specialized algorithms to support machine learning. So-called generative AI (Chat GPT, Dall-E3, Adobe Firefly, Stable Diffusion, etc.) is a subfield of artificial intelligence. According to Kreutzer, there are various types of learning: supervised learning, unsupervised learning, reinforcement learning, and self-supervised learning. He also lists five current areas of application for artificial intelligence \cite{kreutzer}(p. 30):

\begin{itemize}
\item Natural Language Processing (NLP)
\item Natural Image Processing (Computer Vision)
\item Data Collection, Storage, and Information Processing (Expert Systems)
\item Computer-aided Mechanical Systems (Robotics)
\item Affective Artificial Intelligence (Affective AI) 
\end{itemize}

\section{Sustainable Software Development and Evolutionary Design}

Evolutionary design is an approach to sustainable software development in which a product or system is developed incrementally and iteratively - similar to biological evolution. The core idea is that a system is continuously improved over time and adapted to changing conditions. Even today, there are various terms associated with evolutionary design, such as evolutionary and co-evolutionary software development, clean code, green IT, and sustainable AI. Today there are diverse areas of application.

\subsection{Induction Design and Bionics}

For example, the japanese architect Makoto Sei Watanabe developed what is known as ``Induction Design,'' a method of evolutionary design specifically for architectural design \cite{watanabe}. This is an early, algorithm-based, generative method for intelligent form-finding. \\

The architect Frei Otto explored the apparent contrast between the artificial and the natural \cite{otto}. He sees humans themselves and all their technical products as part of nature. In the creation of artificial and technical objects, he focuses on the observation of nature and natural processes. In his designs, he employs methods of bionics. Bionics is an engineering discipline that utilizes natural phenomena to develop technical objects in a resource-efficient manner \cite{rechenberg}. Frei Otto focuses on natural self-forming and optimization processes. Based on the insights gained, he develops resource-efficient, minimalist architecture that is optimized for function and cost. \\

However, these design processes often unfold in surprising and non-linear ways. The starting point is not a deliberate conception of a specific building form (“giving form” or “wanting form”). Only through an evolutionary and experimental design process, the optimal form of the building is discovered iteratively and step by step (“finding form”). \\

\subsection{Evolutionary and Co-Evolutionary Software Development}

Evolutionary and co-evolutionary software development aims to understand the mechanisms of natural evolutionary and co-evolutionary processes in order to gain insights for the development of computer-aided problem-solving systems \cite{chong}. Similar processes and technologies are already being successfully applied in industry, such as in multi-criteria optimization in vehicle development. \\

\subsection{Clean Code, Green Coding, and Green IT}

Martin describes five basic principles of evolutionary design in connection with the concept of clean code \cite{martin}: \\

\begin{itemize}
\item Iterative development
\item Adaptability (evolvability)
\item Refactoring
\item Feedback loops
\item Minimum Viable Product (MVP) \\
\end{itemize}

The development process proceeds in small steps (e.g., in agile sprints or iterations). After each iteration, the system is tested, evaluated, and improved. The design remains open to changes. Requirements may change - and the system adapts without requiring the entire design to be rewritten. Regular “cleanup” and improvement of the existing code or design, without altering the functionality. This ensures that the system does not become “obsolete” or “entangled.” User feedback, test results, and technical insights are continuously incorporated into further development. The process begins with a simple, functional core system and expands it step by step. 

At the heart of Green Coding and Green IT lies the entire life cycle of a software product and its integration into the business environment. This field encompasses the resource-efficient use of energy and environmental resources through information technology. Green IT considers all measures and strategies related to software development and operation across the entire infrastructure, including data centers and end devices. \\

Lieder lists seven guidelines for Green Coding \cite{lieder}: \\

\begin{itemize}
\item Data-driven life cycle analysis
\item Caching and data minimization
\item Carbon awareness
\item Mobile first (runtime-efficient solutions)
\item Efficient algorithms and data structures
\item Avoid over-engineering and obsolescence
\item Keep it simple
\end{itemize}

\subsection{Sustainable AI}

Sonnet et al. identify both negative and positive impacts of artificial intelligence (AI) on social and environmental sustainability (environmental protection, equal treatment, and human rights) \cite{sonnet}. Negative environmental impacts of AI include its enormous carbon footprint and water consumption. Conversely, however, AI can also have positive effects, such as improved problem-solving capabilities in climate research, environmental and process regulation, or the prevention of poverty and hunger. The authors also view AI positively as an interdisciplinary field that brings together many different disciplines in problem-solving. In summary, the authors propose three technological solutions for reducing AI’s energy consumption: \\

\begin{itemize}
\item Knowledge distillation using pre-trained teacher and student models
\item Compact network architectures (MobileNet)
\item Memory optimization through quantization and binarization \\
\end{itemize}

Sonnet also refers to the optimal alignment of AI with the desired application in order to minimize energy and resource consumption. It is precisely this specialization of applications that will become particularly interesting later on, in terms of open and quality-oriented technology development. 

\subsection{Digital Ethics and Ethically Aligned Design}

The aforementioned authors also emphasize the interconnection between AI ethics and sustainability. While sustainability focuses on the needs and opportunities of future generations, ethics calls on us to take responsibility for our actions and for the common good. Many companies today practice Corporate Social Responsibility (CSR) and Corporate Digital Responsibility (CDR). \\

Digital ethics has established itself as a subfield of ethics since 2009. It combines aspects of information ethics, media ethics, and machine ethics. Machine ethics attributes moral agency to machines, even though they cannot assume responsibility. Digital ethics formulates rules for action in conflict situations arising from digitalization and addresses legal frameworks and voluntary standards. Examples of these standards include Brad Smith’s six principles and initiatives such as the “Ethically Aligned Design” guidelines developed by the Institute of Electrical and Electronics Engineers (IEEE). The six core principles are \cite{smith1}\cite{smith2}: \\

\begin{itemize}
\item fairness
\item reliability
\item privacy and security
\item inclusiveness
\item transparency
\item accountability \\
\end{itemize}

In addition, there are further legal regulations at the national and international levels (e.g., the EU’s 2024 AI Act) and various internal corporate frameworks for digital ethics (e.g., Google DeepMind, Microsoft Responsible AI, IBM Cognitive Ethics, and AI Fairness 360). \\

However, this rule-based ethics corresponds more to a deterministic or probabilistic approach and is thus fundamentally different from reflective or reason-based, free behavior. Upon examining these current developments, parallels to the aforementioned human intentions, beliefs, and desires become apparent. The energy and resource consumption of AI suddenly appears comparable to human needs or desires. The intended purpose or area of application of AI can, in principle, correspond to human intentions. In this context, human beliefs correspond to a deterministic or probabilistic ethics of AI.

\section{The Margin of Indeterminacy and the Open Machine}

In his book ``On the Mode of Existence of Technical Objects'', Gilbert Simondon argues against a purely economically determined and highly specialized automaton \cite{simondon1}. He describes the robot as a “mythical and imaginary being” and advocates for a “margin of indeterminacy” in the development of technical products. In his view, specialization serves only the mythically exaggerated notion of full automation in the interest of purely economic goals. Restricting this margin of indeterminacy therefore leads to a misunderstood form of functional specialization and, consequently, to limited opportunities for further technical development. \\

Simondon therefore calls for the “open machine.” He advocates a concept of open function that promises technical advantages for further evolutionary development. In doing so, he describes the openness of meaning inherent in technical objects. This functional openness fosters emergence in technical evolution. He speaks of a certain “margin of indeterminacy” that fosters the coherence of machine ensembles and ensures the “best possible exchange of information” between humans and machines. When discussing the use of technical systems, Simondon describes unpredictable “interactions” between the system components. \\

„ ... and effects occur that are independent of the intention behind their creation“. \cite{simondon1} \\

Simondon thus describes the significance of chance and the existence of unpredictable, emergent processes in the operation of the technical system. Openness thus becomes an opportunity for chance and unpredictable emergent processes that can foster positive technological development. \\

“... the technical object is never fully known; for this very reason, it is never fully concrete, except through an extremely rare coincidence triggered by chance.” \cite{simondon1}

\section{Deterministic Digital Ethics and Open Technological Evolution}

\begin{figure*}[htbp]
\begin{minipage}[b]{1.0\textwidth}
\centerline{\includegraphics[width=0.95\textwidth]{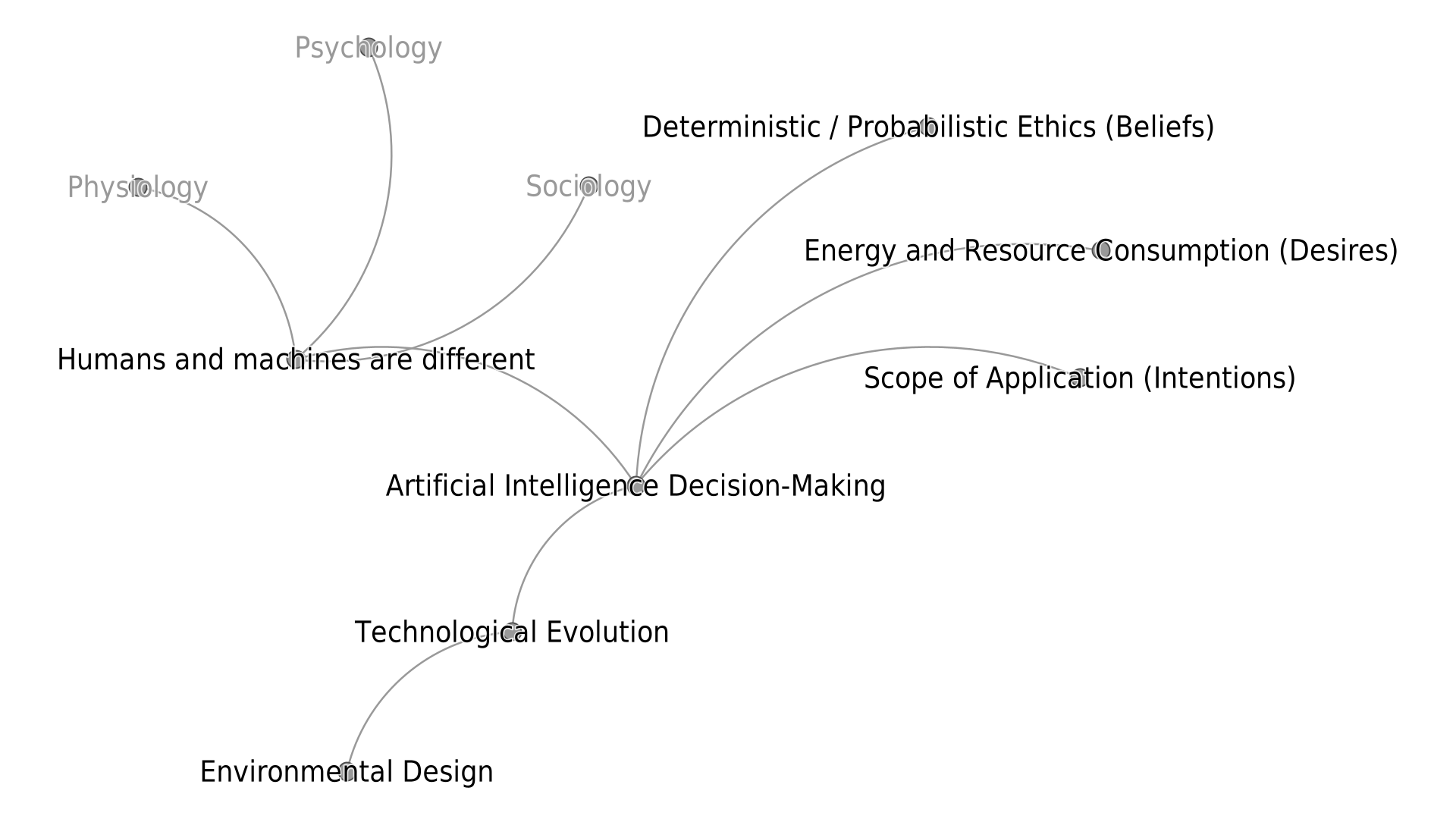}}
\caption{Technological Evolution and Environmental Design as a Result of Artificial Intelligence Decision-Making}
\label{fig:5}
\end{minipage}
\end{figure*}

\subsection{Autonomy and Determinism}

According to Nida-Rümelin and Weidenfeld, it is the problem of autonomy and determinism, of freedom and responsibility, that also constitutes a key difference between Digital Humanism and Evolutionary Design \cite{ruemelin1}. The second key difference is a rule-based ethics in artificial intelligence as opposed to an autonomous ethics of human decision-making. \\

In the following Fig. \ref{fig:5}, technological evolution is interpreted from the perspective of digital humanism in the sense of evolutionary design. For example, human beliefs are assigned to a deterministic or probabilistic digital ethics and corresponding standards for artificial intelligence. Human desires can be mapped to the needs or desires of an artificial intelligence, such as energy or resource consumption. Human intentions can be viewed as the intended purpose or scope of application of a software application.

\subsection{Conviviality and Subjectivity}

Humans and machines are fundamentally different in terms of their “form of life.” Humans and machines do not share a common form of life or existence (conviviality). The machine possesses its own “physiology,” “psychology,” or “sociology” - concepts that are, as of now, largely unimaginable to us. For artificial intelligence, therefore, a genuine subjectivity of a “conversation partner” is fundamentally impossible. Furthermore, AI lacks the capacity for empathic and semantic understanding. AI also lacks intentional and phenomenal consciousness. Consequently, artificial systems are often recognizable by their contextual and phenomenological errors. \\

One goal of future research could be to investigate the “physiology,” “psychology,” or “sociology” of machines. It would be reasonable to expect that artificial networked systems might have fundamentally different foundations than human societies. But wouldn’t this comparative approach be worth a try in order to better understand the foundations of artificial systems?

\subsection{Functional Specialization and Open Technological Evolution}

The precise scope of application of an artificial system has implications for its further technological development. Just as Simondon calls for the “open machine,” digital humanism takes a critical look at purely market- and consumption-oriented technological evolution. Nida-Rümelin also points out sociopolitical interdependencies. Our societal model largely determines how knowledge and data are handled. As an example he cites the term “knowledge society” in contrast to the existing “data economy.” In fact, the current market- and consumption-oriented approach - which favors highly specialized developments rather than further promoting and strengthening quality-oriented ones - poses problems for the open technological development advocated by Simondon. \\

Sustainable AI aims to operate in a resource-conserving and energy-efficient manner, thereby accelerating the development of highly specialized software applications. However, this optimization of sustainable AI for a specific purpose has a negative impact on open and quality-oriented technological development. The freedom of technological evolution is restricted. \\

\begin{figure*}[htbp]
\begin{minipage}[b]{1.0\textwidth}
\centerline{\includegraphics[width=0.95\textwidth]{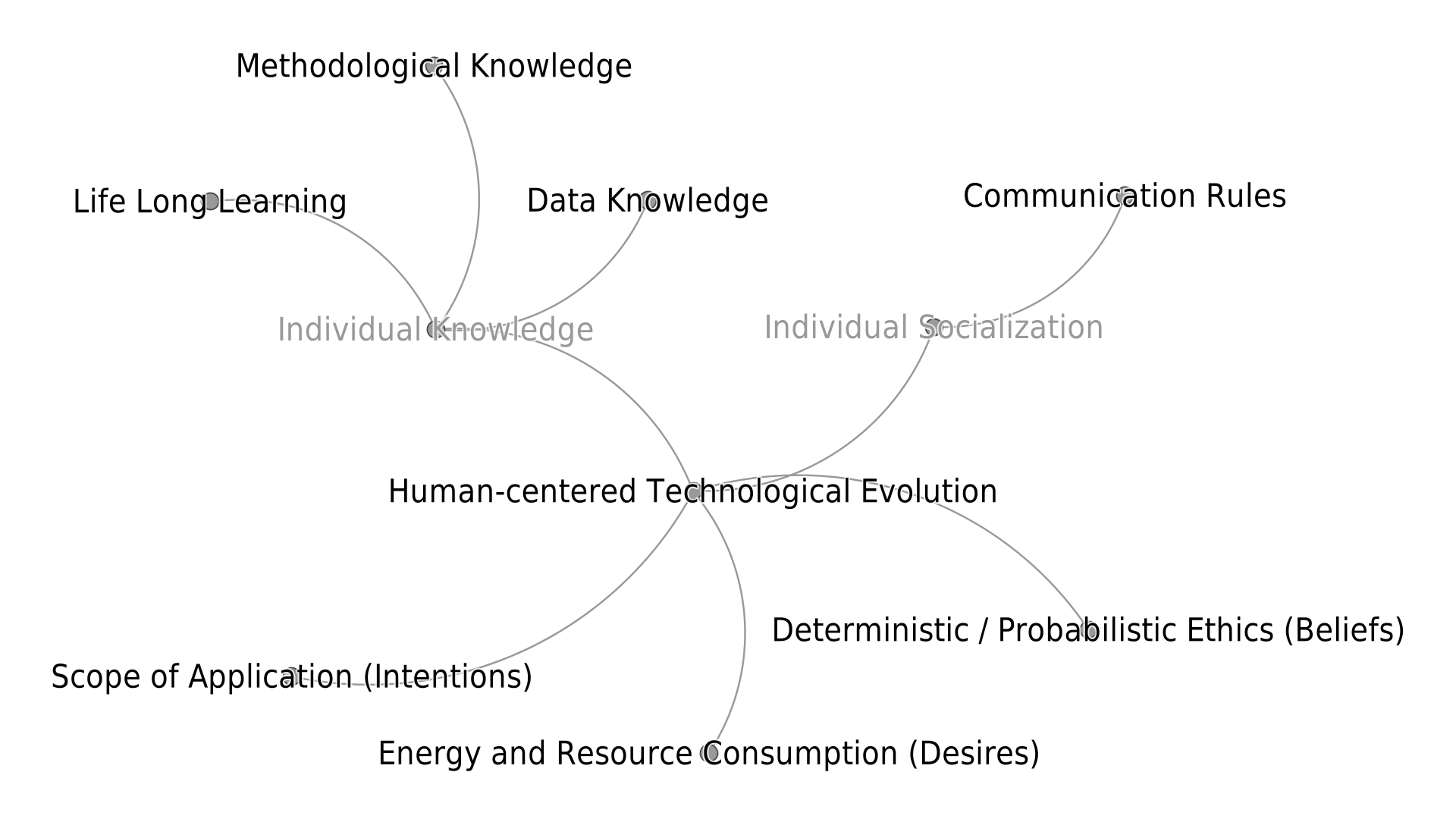}}
\caption{Scope of Application, Energy and Resource Consumption and Ethics as a Result of Human-centered Technological Evolution}
\label{fig:6}
\end{minipage}
\end{figure*}

Furthermore, individual knowledge and socialization are fundamentally different within the framework of humanistic personality development than they are in machine learning models. Fig. \ref{fig:6} illustrates the key differences. While machine learning can very effectively model both data and methodological knowledge, as well as continuous learning and communication rules, two aspects remain poorly understood today: the representation of individual knowledge and individual socialization in an artificial machine. \\

\newpage
\subsection{Human-centered Technological Evolution and an Applied Ontology of Artificial Intelligence}

The concept of digital humanism can be successfully applied to open technological evolution. Digital humanism thus serves more than just humanity. If artificial intelligence or other technologies are to be oriented toward humans, they must also be able to develop technologically in a manner “appropriate to their nature,” with open-ended outcomes, and freely. \\

Analogous to digital humanism, this work therefore calls for a free and open “human-centered technological evolution.” In the spirit of Simondon, this work advocates for an “open machine” that will continue to exist in the future - one that can develop freely and with an open-ended outcome, both in terms of its nature and technologically, within a certain margin of indeterminacy. What is therefore necessary for such a humanistic technological evolution is not, first and foremost, the economic benefit of a new technology, but rather controlled, quality-oriented, and open technological development. \\

This experimental development can only be successful under scientific and controlled conditions before the product reaches the end consumer, as required by the national regulatory sandboxes under the EU’s AI Act \cite{EU}. \\

This work stimulates a new scientific discourse with the goal of better understanding and further exploring the nature of technical objects - an ontology of artificial intelligence - from an application-oriented perspective.


\newpage

\bibliographystyle{ACM-Reference-Format}
\bibliography{sample-base}






\end{document}